\documentclass{ifacconf}

\usepackage{graphicx}      
\usepackage{natbib}        

\usepackage{ifpdf}
\usepackage{lipsum}
\usepackage{color}
\usepackage{graphicx}
\usepackage{amsmath}
\usepackage{amssymb}
\usepackage{listings}
\usepackage[nonumberlist,acronym]{glossaries}
\usepackage{array}
\usepackage{placeins}
\usepackage{stfloats}
\usepackage{url}
\usepackage{multirow}
\usepackage{accents}

\usepackage{float}
\usepackage[table,xcdraw]{xcolor}
\usepackage{multicol,lipsum}
\usepackage{amsfonts}
\usepackage{flushend}

\usepackage[normalem]{ulem} 

\usepackage{accents}  
\newcommand\munderbar[1]{%
	\underaccent{\bar}{#1}}

\DeclareMathOperator{\Tr}{Tr}

\DeclareMathOperator*{\argmin}{arg\,min}

\newcolumntype{P}[1]{>{\centering\arraybackslash}p{#1}}

\usepackage{ctable} 

\graphicspath{{./Figures/}}

\newacronym{ROS}{ROS}{Robot Operating System}

\usepackage{ctable} 

\graphicspath{{./Figures/}}

\definecolor{red}{rgb}{1,0,0} 
\definecolor{blue}{rgb}{0,0,1} 
\definecolor{forestgreen}{rgb}{0.133333,0.545098,0.133333} 
\definecolor{magenta}{rgb}{1,0,1} 
\definecolor{firebrick}{rgb}{0.698039,0.133333,0.133333} 
\definecolor{darkgreen}{rgb}{0,0.392157,0} 
\definecolor{green}{rgb}{0,1,0} 
\definecolor{purple}{rgb}{0.627451,0.12549,0.941176} 
\definecolor{darkcyan}{rgb}{0,0.545098 ,0.545098} 
\definecolor{goldenrod}{rgb}{0.854902,0.647059 ,0.12549} 

\definecolor{darkblue}{rgb}{0,0,0.7}

\definecolor{blue}{rgb}{0,0,1} 
\definecolor{red}{rgb}{1,0,0} 
\definecolor{darkgreen}{rgb}{0,0.392157,0}

\begin{document}
\begin{frontmatter}

\title{End-Effector Stabilization of a 10-DOF Mobile Manipulator using Nonlinear Model Predictive Control}


\author[First]{Mostafa Osman} 
\author[First]{Mohamed W. Mehrez} 
\author[First]{Shiyi Yang}
\author[First]{Soo Jeon}
\author[First]{William Melek}

\address[First]{Mechanical and Mechatronics Engineering, University of Waterloo, Waterloo, Ontario, N2L 3G1, Canada
	(email: \{meaosman, mohamed.said, s268yang, soojeon, william.melek\}@uwaterloo.ca)}

\begin{abstract}                
Motion control of mobile manipulators (a robotic arm mounted on a mobile base) can be challenging for complex tasks such as material and package handling. 
In this paper, a task-space stabilization controller based on Nonlinear Model Predictive Control (NMPC) is designed and implemented to a 10 Degrees of Freedom (DOF) mobile manipulator which consists of a 7-DOF robotic arm and a 3-DOF mobile base. The system model is based on kinematic models where the end-effector orientation is parameterized directly by a rotation matrix.
The state and control constraints as well as singularity constraints are explicitly included in the NMPC formulation. The controller is tested using real-time simulations, which demonstrate 
high positioning accuracy with tractable computational cost.
\end{abstract}


\end{frontmatter}

\section{Introduction}
\label{sec:Introduction}
\begin{figure}[t]
	\centering
	\vspace*{7mm}
	\begin{tabular}{cc}
		\includegraphics [trim=0mm 0mm 0mm 0mm,scale=0.1965]{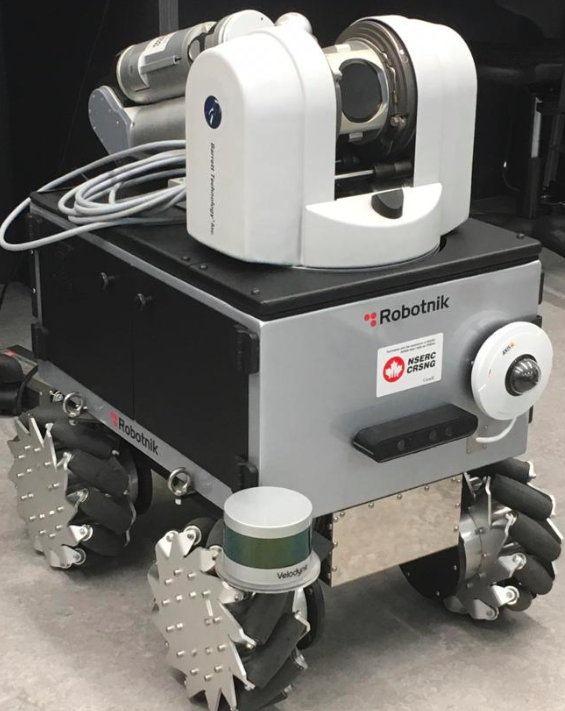} &
		\includegraphics[width=0.35\linewidth]{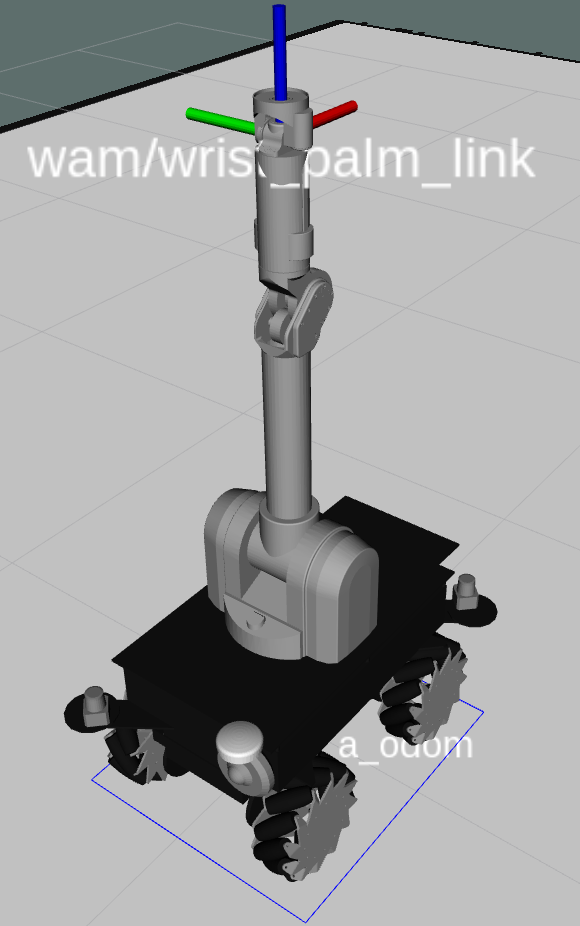} \\
	\end{tabular}
	\caption{\footnotesize The synthesized mobile manipulator. Left: the real robot. Right: the simulated robot.}
	\label{fig:robot}
\end{figure}

The use of mobile manipulators in industry has increased drastically over the past decade. 
Mobile manipulators combine the advantages of both wheeled robots and robotic arms; thus, they have an expandable workspace and operational versatility through perception, object manipulation, and mobility. Such robots can be used in material handling, wall painting, as well as inspection and repairs, see, e.g.~\citep{bostelman2016survey} for a survey. Operating such systems requires safe navigation in possibly dynamic environments and precise object manipulation. In this paper, we focus on stabilizing the end-effector of a 10-DOF mobile manipulator shown in Fig.~\ref{fig:robot}. The mobile manipulator is built by integrating the Summit-XLS mobile robot with meccanum wheels manufactured by Robotnik and the 7-DOF Barrett WAM robotic arm.


The separate control of mobile robots and robotic arms is studied extensively in the literature. The considered control problems can be categorized under point-stabilization, trajectory tracking, and path following. The commonly used control techniques include feedback linearization~\citep{d1995control}, robust control~\citep{koubaa2013robust}, fuzzy based feedback linearization~\citep{piltan2013design}, adaptive control~\citep{pourboghrat2002adaptive,slotine1988adaptive}, and model predictive control (MPC)~\citep{faulwasser2016implementation, Mehrez_Holonomic_2020}. 


Several studies that consider controlling mobile manipulators as a combined system also exist. For example,~\cite{silva2016whole} designed a whole-body controller based on feedback linearization controlling the end-effector pose of a mobile manipulator. The controller was tested on a 7-DOF mobile manipulator. \cite{patel2017adaptive} proposed an adaptive backstepping control for the trajectory tracking of mobile manipulators. \cite{mishra2018robust} developed a robust nonlinear controller with uncertainty estimator; the controller was validated through simulations for a 7-DOF mobile manipulator. Furthermore, \cite{avanzini2016reactive} used linear MPC for developing a reactive constrained controller of an omnidirectional mobile base with a 5-DOF robotic arm. All the aforementioned studies considered mobile manipulators with non-redundant arms. This simplifies the problem due to the presence of a closed-form inverse kinematics solutions for such arms. 
Thus, the control of the mobile manipulator end-effector can be designed in the configuration (joint) space by utilizing a separate closed-form inverse kinematics module, see, e.g.~\citep{avanzini2016reactive}.


MPC is popular in the field of controls because of its ability to handle constrained possibly nonlinear mutli-input-multi-output (MIMO) systems. In MPC, a cost function characterizing the control objective is minimized using an open-loop control sequence or function while state and control constraints are considered. The first part of the resulting open-loop control is then applied to the system. Finally, the process is repeated every decision instant, see, e.g.~\citep{allgower2012nonlinear}.

In this paper, we use a nonlinear model predictive control (NMPC) scheme to stabilize the end-effector of a 10-DOF mobile manipulator; 
here, we first formulate the task-space kinematic model, where the overall system rotations are expressed using the 3D special orthogonal $SO(3)$ representation. Then, this model is used for state prediction in the NMPC formulation, which considers state and control constraints as well as kinematic singularity constraints. 
The proposed NMPC controller is implemented using Robot Operating System (ROS)~\citep{Quigley09} and the efficacy of the proposed controller is demonstrated through a series of real-time  simulations using Gazebo dynamic simulator~\citep{koenig2004design}. The results show highly-accurate and smooth stabilization of the end-effector as well as computational cost, which meets the real-time requirements.


The remainder of the paper is organized as follows: the model used to implement the NMPC is explained in Section~\ref{sec:modeling} followed by the optimal control (OCP) problem formulation in Section~\ref{sec:MPC}. In Section~\ref{sec:simulations}, the simulation testbed used to validate the proposed controller is introduced and then the acquired results are shown in Section~\ref{sec:results_and_discussion}. Finally, in Section~\ref{sec:conclusion} conclusions are stated and the future work is summarized.

\section{Mobile Manipulator Modeling}
\label{sec:modeling}
In this section, we, first, show the notations used in the paper. Then, we present the kinematic model of the synthesized mobile manipulator.

\subsection{Notations}
$\mathbb{N}$ and $\mathbb{R}$ denote the sets of natural and real numbers, respectively, $||x||_{\infty}$ is the $l_{\infty}$ norm defined as $||x||_{\infty} := \max_{i\in[1:n]} |x_{i}|$, $||x||^{2}_{A}$ is the squared $l_{2}$ norm weighted by $A$ and is calculated as $x^{\top}Ax$, the matrix trace operator is denoted by $\Tr$, $A \succ 0$ denotes that $A$ is a positive definite matrix, $\mathbf{I}_{n\times n}$ is the $n\times n$ identity matrix, $\mathbf{0}_{n \times k}$ denotes an $n\times k$ matrix with all entries of zeros, and $SO(3)$ and $SE(3)$ denote the special orthogonal and special Euclidean groups, respectively.

\subsection{Kinematic Model of the holonomic Mobile Base}
The mobile base of the considered mobile manipulator is a holonomic mobile robot with meccanum wheels. Holonomic mobile robots possess an extra degree of maneuverability when compared to the non-holonomic robots~\citep{siegwart2011introduction}.

The discrete-time kinematic model of the mobile base is given by 
\begin{align} \label{equ:kinematic_model}
\mathbf{x}^{b}_{k+1} &= \mathbf{f}_{mr}(\mathbf{x}^{b}_{k}, \mathbf{u}_{k}) \nonumber  \\ 
&= \begin{bmatrix}
x^{b}_{k} \\ y^{b}_{k} \\ \gamma_{k}
\end{bmatrix} +\tau  \underbrace{\begin{bmatrix}
	\cos \gamma_{k} & -\sin \gamma_{k} & 0 \\
	\sin \gamma_{k} & \cos \gamma_{k} & 0 \\
	0 & 0 & 1 \\
	\end{bmatrix}}_{R^{I}_{b}} \begin{bmatrix}
u_{1,k} \\  u_{2,k} \\  u_{3,k}
\end{bmatrix},
\end{align}
where $\mathbf{x}^{b} = (x, y, \gamma)^{\top} \in X_{mr} \subset \mathbb{R}^{3}$ is the pose of the mobile base in the inertial frame, $(x^{b}, y^{b})$ are the two Cartesian planar coordinates and $\gamma$ is the yaw angle of the mobile base.
$\mathbf{u} = (u_{1}, u_{2}, u_{3})^{\top} \in U_{mr} \subset \mathbb{R}^{3}$ is the robot input speeds,
\noindent
$\mathbf{f}_{mr}:\mathbb{R}^{3} \times \mathbb{R}^{3} \rightarrow \mathbb{R}^{3}$ is a nonlinear mapping, $R^{I}_{b} \in SO(3)$ is the $z$-axis rotation matrix, and $\tau > 0$ is the sampling time. 

The state constraint set $X_{mr}$ is a compact set defined as 
\begin{equation*}
X_{mr} := [\munderbar{x}^{b}, \bar{x}^{b}] \times [\munderbar{y}^{b}, \bar{y}^{b}] \times [-\pi, \pi], 
\end{equation*}
\noindent where $\munderbar{x}^{b}, \munderbar{y}^{b}, \bar{x}^{b}, \bar{y}^{b}$ are the lower and upper bounds of the Cartesian coordinates $x^{b}$ and $y^{b}$, respectively.

The relation between the robot input speeds $\mathbf{u}$ and wheel speeds $\mathcal{V} = (\omega_{1}, \omega_{2}, \omega_{3}, \omega_{4})^{\top}$, for meccanum wheels robots, can be stated as ~\citep{Lynch:2017:MRM:3165183} 
\begin{equation}
\mathcal{V} := \begin{bmatrix}
\omega_{1} \\ \omega_{2} \\ \omega_{3} \\ \omega_{4}
\end{bmatrix}= H \mathbf{u} = \frac{1}{r} \begin{bmatrix}
1 & -1 & -l-w \\
1 & 1 & l+w \\
1 & -1 & l+w \\
1 & 1 & -l-w \\
\end{bmatrix}\mathbf{u},
\end{equation}
\noindent
where $r$ is the mecanum wheel radius, $l$ and $w$ are half of the wheelbase and the trackwidth, respectively.
Consequently, the control constraint set $U_{mr}$ is a compact set defined as
\begin{equation}
U_{mr} := \{ \mathbf{u} \in \mathbb{R}^{3} | \ ||H \mathbf{u}||_{\infty} \leq \omega_{max}  \},
\end{equation}
\noindent
where $\omega_{max}$ is the rated speed of the wheel motors.

\subsection{Kinematic Model of the Robotic Arm}
The robotic arm mounted on the aforementioned mobile base is a 7-DOF WAM arm by Barrett Technology~\citep{WAM2018}. The end-effector pose of the robotic arm is denoted by $\mathbf{x}^{a} := [\mathbf{p}^{a\top}, \mathbf{\theta}^{a\top}]^{\top} $, where $\mathbf{p}^{a} \in \mathbb{R}^{3}$ is the end-effector position in the robotic arm base frame represented in the Cartesian coordinates and $\mathbf{\theta}^{a}$ is the end-effector orientation. The end-effector orientation can be represented using several methods as discussed in~\citep{campa2009pose}. Here, we use the $SO(3)$ group to avoid representation singularities and error definition discontinuities. To do so, we define the mapping function $f: SO(3) \rightarrow \mathcal{F} \subset \mathbb{R}^{9}$, such that, for a rotation matrix $R \in SO(3)$,
\begin{equation}\label{equ:theta}
f(R) = \begin{bmatrix}
[R]_{1}^\top & [R]_{2}^\top & [R]_{3}^\top
\end{bmatrix}^{\top},
\end{equation}
where $[R]_{i}$, $i \in \{1,2,3\}$, is the $i$-th column vector of the rotation matrix $R$. Thus, the orientation vector $\theta^a$ is $\theta^{a} = f(R_{E}^{b}) \in \mathcal{F}$, and $R^{b}_{E}$ is the rotation matrix of the end-effector in the base frame of the robotic arm.

Using such a representation, the kinematic model of the robotic arm can be described using the analytical Jacobian $\mathbf{J}_{a}$ of the forward kinematics transformation matrix $\mathbf{T} \in SE(3)$ derived using the DH-parameters of the robotic arm, (see~\cite{WAMmanual2018} for the DH-parameters of the considered robotic arm), as 
\begin{equation}
\label{equ:robot_arm_model_so3}
\mathbf{x}^{a}_{k+1} = \mathbf{f}_{ra}(\mathbf{x}^{a}_{k},\mathbf{q}_{k}, \dot{\mathbf{q}}_{k}) = \mathbf{x}^{a}_{k} + \tau \mathbf{J}_{a}(\mathbf{q}_{k})\dot{\mathbf{q}}_{k},
\end{equation}
where $\mathbf{x}^{a} = [\mathbf{p}^{a\top}, \theta^{a\top}]^{\top} \in X_{ra} \subset \mathbb{R}^{12}$ is the state vector defined using $\theta^{a}$ from \eqref{equ:theta}, $\mathbf{q} = [q_{1}, q_{2}, q_{3}, q_{4}, q_{5}, q_{6}, q_{7}]^{\top} \in Q \subset \mathbb{R}^{7}$ is the joint angles vector, $\dot{\mathbf{q}} \in \Omega \subset \mathbb{R}^{7}$ is the joint velocities vector, and, the analytical Jacobian $\mathbf{J}_{a}$ is given by $\mathbf{J}_{a} := \partial \mathbf{T}/\partial \mathbf{q}$.

The constraint sets for the end-effector $X_{ra}$, joint angles $Q$, and joint velocities $\Omega$ are defined by
\begin{align*}
	 X_{ra} &:= [\munderbar{x}^{a}, \bar{x}^{a}] \times [\munderbar{y}^{a}, \bar{y}^{a}] \times [\munderbar{z}^{a}, \bar{z}^{a}] \times \mathcal{F}, \\
	 Q &:= \{ \mathbf{q} \in \mathbb{R}^{7} | \munderbar{q}_{i} \leq q_{i} \leq \bar{q}_{i}, \forall i \in \{1, ..., 7\} \}, \\
	 \Omega &:= \{\dot{\mathbf{q}} \in \mathbb{R}^{7} | \ ||\dot{\mathbf{q}}||_{\infty} \leq \dot{q}_{max}\},
\end{align*}
where $\munderbar{q}_{i}$ and $\bar{q}_{i}$ denote the lower and upper limits of the joint angles, respectively. 

In order to be able to keep track of the joint angles and consider joint constraints, we extend system~\eqref{equ:robot_arm_model_so3} to
\begin{equation}
\label{equ:conc_model}
\begin{bmatrix}
\mathbf{x}^{a}_{k+1} \\  \mathbf{q}_{k+1}
\end{bmatrix} = \begin{bmatrix}
\mathbf{x}^{a}_{k} \\  \mathbf{q}_{k}
\end{bmatrix} + \tau \begin{bmatrix}
\mathbf{J}_{a}(\mathbf{q}) \\ I_{7 \times 7}
\end{bmatrix}\dot{\mathbf{q}}_{k},
\end{equation}
\noindent
where $[\mathbf{x}^{a\top}, \mathbf{q}^{\top}]^{\top} \in \bar{X}_{ra} \subset \mathbb{R}^{19}$ is the concatenated state vector, and $\bar{X}_{ra}$ is the state constraint set for the new augmented model and is defined as $\bar{X}_{ra} := X_{ra} \times Q$.

\subsection{Mobile Manipulator Kinematic Model}
The model of the mobile manipulator can now be derived using the model of the mobile base and the robotic arm. 
First, we map the velocity components of the mobile base to the end-effector linear speeds in the inertial frame, i.e. we have 
\begin{equation}
\label{equ:mr_position}
\dot{\mathbf{p}}^{b} = [R^{I}_{b}]_{2 \times 2} \begin{bmatrix}
u_1  \\ u_{2}   \\
\end{bmatrix} + \underbrace{[R^{I}_{b}]_{2 \times 2} \begin{bmatrix}
-y^{b}_{E} \\ x^{b}_{E}
\end{bmatrix}}_{\psi} u_{3}
\end{equation}
where $\dot{\mathbf{p}}^{b} = [\dot{x}^b, \dot{y}^b]$ is the linear velocity of the end-effector caused by the mobile base in the inertial frame, and $[R^{I}_{b}]_{2 \times 2}$ is the upper left $2 \times 2$ sub-matrix of $R^{I}_{b}$ shown in Eq.~\eqref{equ:kinematic_model}. Note that $R^{I}_{b}$ is the orientation of the mobile robot, which can be determined through the localization feedback. $x^{b}_{E}, y^{b}_{E}$ are the position of the end-effector in the mobile base frame and can be determined from the forward kinematics transformation matrix $\mathbf{T}$.


Second, to calculate the angular velocity of the mobile robot in $SO(3)$, we need to calculate the derivative of the rotation matrix $R^{I}_{b}$. As mentioned in~\citep{campa2009pose}, the derivative of a rotation matrix $R$ can be calculated as
\begin{equation}
\label{equ:skew_symmetric}
\dot{R} = S(\omega) R,
\end{equation}
\noindent
where $S(\omega)$ is the skew symmetric matrix form of the angular velocity vector $\omega = [\omega_{x}, \omega_{y}, \omega_{z}]^{\top}$, where $\omega_{x}, \omega_{y}$ and $ \omega_{z}$ are the three angular velocities around the principle axes $x, y$ and $z$, respectively. Using the properties of the skew-symmetric matrix, Eq.~\eqref{equ:skew_symmetric} can be written as
\begin{equation} \label{equ:theta_b_dot}
\dot{\theta}^{b} := \begin{bmatrix}
[\dot{R}]_{1} \\ [\dot{R}]_{2} \\ [\dot{R}]_{3}
\end{bmatrix} = -\underbrace{\begin{bmatrix}
S([R]_{1}) \\
S([R]_{2}) \\
S([R]_{3}) \\
\end{bmatrix}}_{=: \Theta \in \mathbb{R}^{9 \times 3}}\omega,
\end{equation}
\noindent
where $\dot{\theta}^{b} $ is the rate of change of the end-effector orientation due to the mobile base rotation. Moreover, $S([R]_{n})$ is the $n$-{th} column vector of $R$ in the skew symmetric form.

Since we do not consider any other angular velocities than $w_z$ for the mobile base, the first two columns of $\Theta$ in~\eqref{equ:theta_b_dot} will be zeros. Consequently, $\dot \theta^{b}$ reads
\begin{equation}
\label{equ:mr_orientation}
\dot \theta^{b} := \begin{bmatrix}
[\dot{R^{I}_{b}}]_{1} \\ [\dot{R^{I}_{b}}]_{2} \\ [\dot{R^{I}_{b}}]_{3}
\end{bmatrix} = -\underbrace{\begin{bmatrix}
S([R^{I}_{b}]_{1}) \\
S([R^{I}_{b}]_{2}) \\
S([R^{I}_{b}]_{3}) \\
\end{bmatrix}_{3}}_{=: \Theta_{3} \in \mathbb{R}^{9 \times 1}} u_{3},
\end{equation}
where we exploit the fact that $w_{z}$ for an omnidirectional mobile robot is the control action $u_{3}$ shown in Eq.~\eqref{equ:kinematic_model}.


Using Eq. \eqref{equ:mr_position} and \eqref{equ:mr_orientation}, in addition to the kinematic model in Eq. \eqref{equ:conc_model}, the kinematic model of the whole mobile manipulator can be written as
\begin{align} \label{equ:full_model}
	\begin{bmatrix}
	\mathbf{x}_{k+1} \\  \mathbf{q}_{k+1}
	\end{bmatrix} &= \begin{bmatrix}
	\mathbf{x}_{k} \\  \mathbf{q}_{k}
	\end{bmatrix} + \tau  \mathbf{J}_{mm}(\mathbf{x}_{k}, \mathbf{q}_{k}) \begin{bmatrix}
	\mathbf{u}_{k} \\ \dot{\mathbf{q}}_{k}
	\end{bmatrix} \\ &=  \underbrace{\begin{bmatrix}
\mathbf{x}_{k} \\  \mathbf{q}_{k}
\end{bmatrix} +  \tau
	\left[
	\begin{array}{c|c}
	\begin{array}{c c c}
	& [R_{b}^{I}]_{2 \times 2} & \begin{array}{c c} \psi \\ 0  \end{array} \\
	& \mathbf{0}_{10 \times 2} & -\Theta_{3}
	\end{array}  &  
	\mathbf{J}_{a}(\mathbf{q}_{k})\\
	\hline
	\mathbf{0}_{7 \times 3} & \mathbf{I}_{7 \times 7}
	\end{array}
	\right]
	\begin{bmatrix}
	\mathbf{u}_{k} \\ \dot{\mathbf{q}}_{k}
	\end{bmatrix}
}_{\mathbf{f}_{mm}(\mathbf{x}_{k}, \mathbf{q}_{k}, \mathbf{u}_{k}, \dot{\mathbf{q}}_{k})}, \nonumber
\end{align}
\noindent
where $[\mathbf{x}^{\top}, \mathbf{q}^{\top}]^{\top} \in X \subset \mathbb{R}^{19}$ is the concatenated state vector of the mobile manipulator, $[\mathbf{u}, \dot{\mathbf{q}}^{\top}]^{\top} \in U \subset \mathbb{R}^{10}$ is the concatenated control vector. Here, $\mathbf{x} := [\mathbf{p}^{\top}, \mathbf{\theta}^{\top}]^{\top}$ is the end-effector pose vector in the inertial frame.

The model stated in \eqref{equ:full_model} is the complete kinematic model of the considered 10-DOF mobile manipulator consisting of a 3-DOF holonomic mobile base and a 7-DOF robotic arm. The constraints over the developed kinematic model can now be defined as
\begin{align} \label{eq:mm_constraints}
	X &:= [\munderbar{x}, \bar{x}] \times [\munderbar{y}, \bar{y}] \times [\munderbar{z}, \bar{z}] \times \mathcal{F} \times Q, \text{ and }  \\
	U &:=  U_{mr} \times \Omega. \nonumber
\end{align}

Finally, the end-effector pose feedback can be determined by the pose of the mobile robot determined through the use of a localization algorithm~\citep{osman2019intelligent} 
and the forward kinematic equations of the robotic arm $\mathbf{T}$.

\section{Nonlinear Model Predictive Control}
\label{sec:MPC}
In this section, we formulate an NMPC scheme for the end-effector pose stabilization of the mobile manipulator. To this end, we define
\begin{align*}
	\mathcal{U}_{N} &:= \left(\begin{bmatrix}
	\mathbf{u}_{k} \\ \dot{\mathbf{q}}_{k}
	\end{bmatrix}, 	\begin{bmatrix}
	\mathbf{u}_{k+1} \\ \dot{\mathbf{q}}_{k+1}
	\end{bmatrix}, \dots,  \begin{bmatrix}
	\mathbf{u}_{k+N-1} \\ \dot{\mathbf{q}}_{k+N-1}
	\end{bmatrix} \right) \text{ and}\\
	\mathcal{X}_{N} &:= (\mathbf{x}_{k}, \mathbf{x}_{k+1}, \dots, \mathbf{x}_{k+N} )
\end{align*}
as the sequences of controls and states over the prediction horizon $N \in \mathbb{N}$, respectively. As standard in NMPC, these sequences are used to form the quadratic cost function
\begin{align} \label{equ:cost}
J(\mathcal{U}_{N}, \mathcal{X}_{N}) &= \underbrace{||\mathcal{E}_{N}^{\mathbf{p}}||^{2}_{\mathcal{S}^{\mathbf{p}}} + ||\mathcal{E}_{N}^{\theta}||^{2}_{\mathcal{S}^{\theta}} }_{J_{f}} \nonumber \\   
&+  \sum_{i = k}^{k+N-1} ||\mathcal{E}^{\mathbf{p}}_{k}||^{2}_{\mathcal{Q}^{\mathbf{p}}} + ||\mathcal{E}^{\theta}_{k}||^{2}_{\mathcal{Q}^{\theta}} + \left\|\begin{matrix}
	\mathbf{u}_{i} \\ \dot{\mathbf{q}}_{i}
	\end{matrix} \right\|^{2}_{\mathcal{R}} ,
\end{align}
\noindent
where $\mathcal{S}^{\mathbf{p}} \in \mathbb{R}^{3\times 3} \succ 0, \mathcal{S}^{\theta} \in \mathbb{R}^{9 \times 9} \succ 0, \mathcal{Q}^{\mathbf{p}} \in \mathbb{R}^{3\times 3} \succ 0, \mathcal{Q}^{\theta} \in \mathbb{R}^{9\times 9} \succ 0 $ and $\mathcal{R} \in  \mathbb{R}^{10 \times 10} \succ 0$ are the weighting matrices of the quadratic cost function, and $J_{f}$ is the terminal cost of the cost function. $\mathcal{E}^{\mathbf{p}} \in \mathbb{R}^{3}$ is the translational error of the end-effector pose defined as $\mathcal{E}^{\mathbf{p}} := \mathbf{p} - \mathbf{p^{r}}$, where $\mathbf{p}^{r}$ is the reference position, and
$\mathcal{E}^{\theta} \in \mathbb{R}^{9}$ is the orientation error of the end-effector pose defined as
\begin{equation}
\label{equ:orientation_error}
\mathcal{E}^{\theta} := \begin{bmatrix}
[I_{3 \times 3}]_{1} \\ [I_{3 \times 3}]_{2} \\ [I_{3 \times 3}]_{3} \\ 
\end{bmatrix} - \begin{bmatrix}
[(R^{I}_{E})^{\top} R_{r}]_{1} \\ [(R^{I}_{E})^{\top} R_{r}]_{2} \\ [(R^{I}_{E})^{\top} R_{r}]_{3}\\
\end{bmatrix},
\end{equation}
where $R_{r}$ is the reference orientation, and $R^{I}_{E}$ is the orientation of the end-effector in the inertial frame calculated as $R^{I}_{E} = R^{I}_{b}R^{b}_{E}$. $R^{I}_{b}$ is determined using a localization algorithm of the mobile robot and $R^{b}_{E}$ is the rotation matrix  from the mobile robot to the end-effector and is calculated from the forward kinematics of the robotic arm, i.e. $\mathbf{T}$.

Using the cost function in Eq.~\eqref{equ:cost}, the NMPC optimal control problem can be formulated as:
\begin{subequations}
	\label{equ:opt_problem_mul}
	\begin{alignat}{2}
	&       & (\mathcal{U}_{N}^{*} ,\mathcal{X}_{N}^{*} ) = & \argmin_{\mathcal{U}_{N} \in U, \mathcal{X}_{N} \in X} J(\mathcal{U}_{N}, \mathcal{X}_{N}) \label{eq:opt}\\
	& \text{subject to} &      & \hspace*{-14mm}
	\begin{bmatrix}
	\mathbf{x}_{k+1} & \mathbf{q}_{k+1}
	\end{bmatrix}^\top - \mathbf{f}_{mm}(\mathbf{x}_{k}, \mathbf{q}_{k}, \textbf{u}_k, \dot{\mathbf{q}}_{k})  = 0
	,\label{eq:constraint1_mul}\\
	&                  &      & \mathcal{X}_{N} \in X \subseteq \mathbb{R}^{19}, \label{eq:constraint2_mul} \\
	&                  &      & \mathcal{U}_{N} \in U \subseteq \mathbb{R}^{10}, \label{eq:constraint3_mul} \\
    &                  &      & |det(\mathbf{J}_{a} \mathbf{J}_{a}^{T})| > \epsilon \label{eq:constraint4_mul} 
	\end{alignat}
\end{subequations}
where $\epsilon$ is a threshold for avoiding singular configurations of the robotic arm.

OCP~\eqref{equ:opt_problem_mul} is converted to a nonlinear programming problem (NLP) using the direct multiple-shooting method~\citep{doi:10.1137/080724885}. Here, both the control sequence  $\mathcal{U}_{N}$ as well as the state sequence  $\mathcal{X}_{N}$ are considered as decision variables in the optimization problem. Moreover, the system model is considered as an optimization constraint as formulated by Eq.~\eqref{eq:constraint1_mul}. Multiple-shooting discretization technique provides a more computationally efficient solution to OCP~\eqref{equ:opt_problem_mul} when compared with other discretization techniques, e.g. single-shooting, see~\citep{doi:10.1137/080724885} for more details. Finally, state and control constrains are considered by means of Eq.~\eqref{eq:constraint2_mul} and \eqref{eq:constraint3_mul}.
Note that the inequality constraint~\eqref{eq:constraint4_mul} is added to avoid kinematic singularities of the robotic arm through operation. This is accomplished through ensuring that the pseudo-inverse of the robot arm Jacobian matrix is always invertible and, thus, singular configurations are avoided.




The feedback control law can now be stated as
\begin{equation*}
\begin{bmatrix}
\mathbf{u}_{k}^{*} & \dot{\mathbf{q}}_{k}^{*}
\end{bmatrix}^\top := \mathcal{U}_{N}^{*}(0),
\end{equation*}
\noindent
i.e. the feedback control is the first element in the optimal control sequence $\mathcal{U}_{N}^{*}$. Moreover, the resulting feedback system can be stated as
\begin{equation*}
\begin{bmatrix}
\mathbf{x}_{k+1} &  \mathbf{q}_{k+1}
\end{bmatrix}^\top = \mathbf{f}_{mm}(\mathbf{x}_{k}, \mathbf{q}_{k}, \mathbf{u}^{*}_{k}, \dot{\mathbf{q}}_{k}^{*}).
\end{equation*}

\section{Real-Time Simulation Testbed and Simulation Scenarios} \label{sec:simulations}
\begin{figure} [tp]
	\centering
	\includegraphics[width=0.7\linewidth]{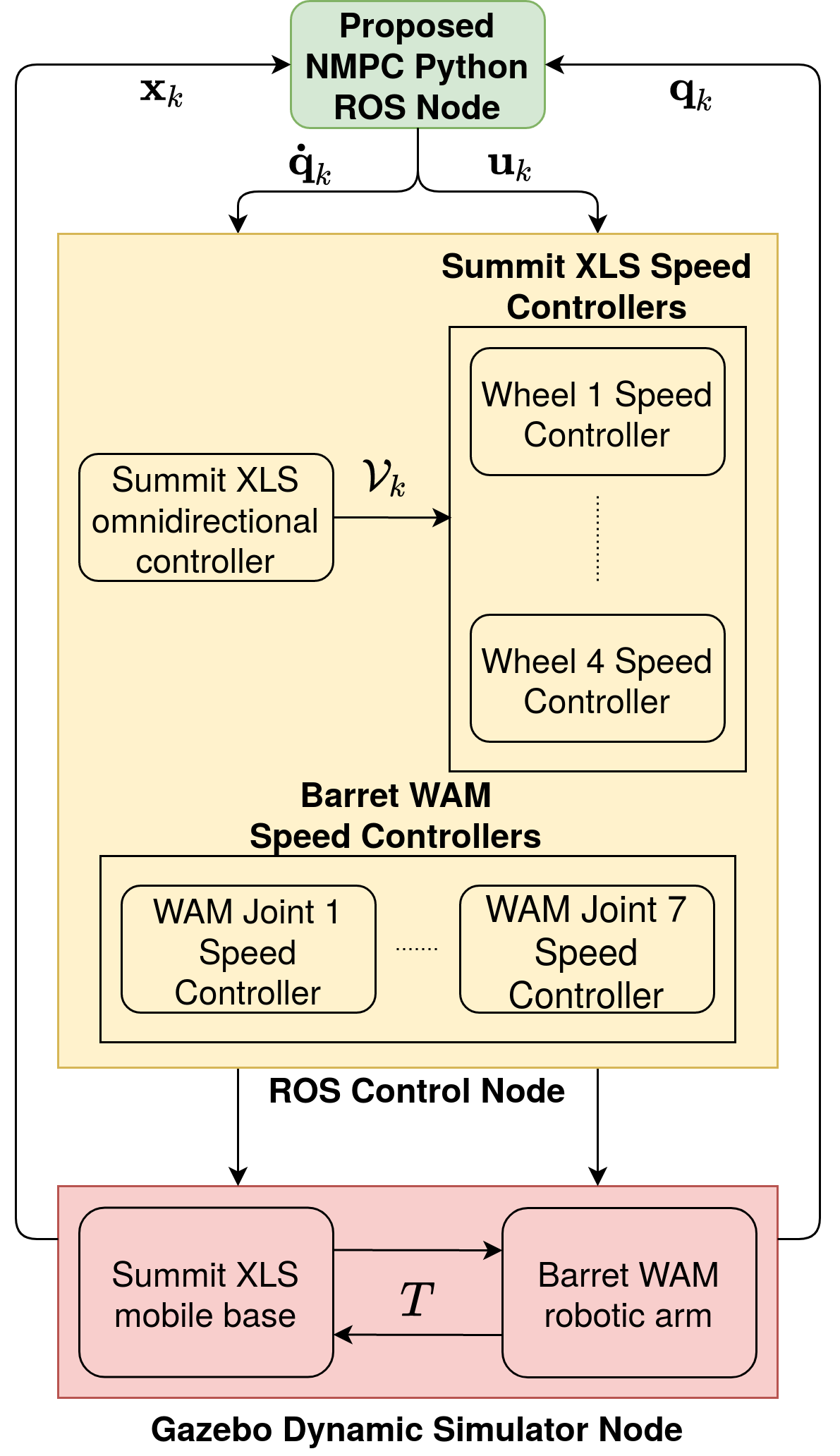}
	\caption{\footnotesize Block-diagram of ROS/Gazebo dynamic simulation environment used to validate the proposed controller.}
	\label{fig:simulation}
\end{figure}

The simulation testbed of the considered mobile manipulator is created by synthesizing the ``urdf" models of both the Barret WAM arm and the Summit XLS mobile robot in a ROS/Gazebo simulation environment\footnote{urdf: universal robotic description format.}. Here, the transformation and the constraints between the two models are defined based on the actual physical system, see Fig.~\ref{fig:robot}. Moreover, the proposed NMPC controller is programmed in python programming language and integrated with the simulation environment via a ROS-node. Here, OCP~\eqref{equ:opt_problem_mul} is formulated symbolically using the numerical optimization software tool CasADi~\citep{Andersson2018}. Additionally, OCP~\eqref{equ:opt_problem_mul} is solved using the interior-point optimization method via the open source solver IPOPT~\citep{Wachter2006}. The overall block-diagram of the ROS/Gazebo real-time simulation environment is illustrated in Fig.~\ref{fig:simulation}.

\begin{table}[bp] 	
	\label{tab:scenarios}
	\caption{Real-time simulation scenarios}
	\centering
	\renewcommand{\arraystretch}{1.3}
	\begin{tabular}{|c|l|l|}
		\hline
		Scenario & Initial States & Reference Pose                       \\ \hline
		1        & $[2, 0, 1.42, 0, 0, 0]^{\top}$ & $[0, 0, 0.5, \pi, 0, 0]^{\top}$   \\ \hline
		2        & $[2, 2, 1.42, 0, 0, 0]^{\top}$ & $[0, 0, 0.5, \pi/2, 0, 0]^{\top}$   \\ \hline
		3        & $[0, 2, 1.42, 0, 0, 0]^{\top}$ & $[0, 0, 0.5, 0, \pi/2, 0]^{\top}$   \\ \hline
		4        & $[-2, 2, 1.42, 0, 0, 0]^{\top}$ & $[0, 0, 0.5, 0, \pi, 0]^{\top}$  \\ \hline
		5        & $[-2, 0, 1.42, 0, 0, 0]^{\top}$ & $[0, 0, 0.5, 0, 0, \pi/2]^{\top}$  \\ \hline
		6        & $[-2, -2, 1.42, 0, 0, 0]^{\top}$ & $[0, 0, 0.5, 0, 0, \pi]^{\top}$  \\ \hline
		7        & $[0, -2, 1.42, 0, 0, 0]^{\top}$ & $[0, 0, 0.5, 0, 0, 0]^{\top}$  \\ \hline
		8        & $[2, -2, 1.42, 0, 0, 0]^{\top}$ & $[0, 0, 0.5, \pi, 0, 0]^{\top}$  \\ \hline
	\end{tabular}
\end{table}

The scenarios shown in Table~\ref{tab:scenarios} were used to validate the proposed controller. Initial and set-point references shown are presented using the Z-Y-X Euler angles to simplify the presentation.
In all scenarios, the sampling time is $\tau = 0.15$ sec, the prediction horizon is $N = 5$, and the weighting matrices are $\mathcal{S}^{\mathbf{p}}, \mathcal{Q}^{\mathbf{p}} = 2\mathbf{I}_{3\times3}$, $\mathcal{S}^{\theta}, \mathcal{Q}^{\theta} = 15\mathbf{I}_{9\times9}$ and $\mathcal{R} = \mathbf{I}_{10\times10}$. 
Furthermore, the constraints set $X$ defined in~\eqref{eq:mm_constraints} is chosen as 
\begin{equation*}
X = [-3, 3] \times [-3, 3] \times [0.4, 1.43] \times \mathcal{F} \times Q,
\end{equation*}
where the arm joint angles limits set $Q$ is given by 
\begin{equation*}
Q = \left\{ \begin{bmatrix}
-2,6 \\ -1.985 \\ -2.8 \\ -0.9 \\ -4.55 \\ -1.5707 \\ -3.0
\end{bmatrix} \leq \mathbf{q} \leq \begin{bmatrix}
2.6 \\ 1.985 \\ 2.8 \\ \pi \\ 1.25 \\ \pi/2 \\ 3.0 \\
\end{bmatrix} \right\}.
\end{equation*}
Here, the joint angle limits are adapted from the arm specifications~\citep{WAM2018}. Finally, the limit of the joints speeds is chosen as $\dot{q}_{max} = 0.5$ rad/sec and the angular speed limit of the mobile robot wheels is chosen as $\omega_{max} = 0.6$ rad/sec.



\section{Results and Discussion} \label{sec:results_and_discussion}
\begin{figure}[bp]
	\centering
	\includegraphics[trim=15mm 10mm 20mm 5mm,scale=0.475]{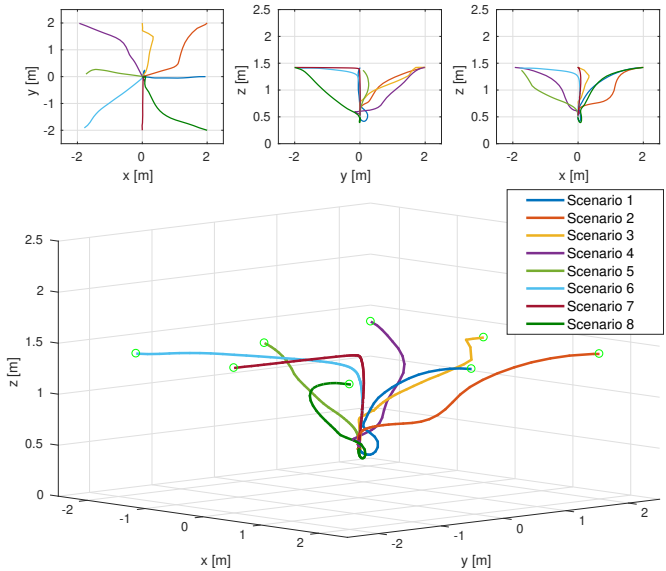}
	\caption{\footnotesize The position trajectories of the end-effector for all the scenarios. Bottom: 3D visualization of the trajectories taken by th end-effector. Top: The three projected views of the end-effector trajectories.}
	\label{fig:scanrios_fig}
\end{figure}

In this section, we show the closed-loop results of the mobile manipulator under the NMPC controller for all the scenarios stated in Table~\ref{tab:scenarios}. Fig.~\ref{fig:scanrios_fig} shows the trajectory of the end-effector through each simulation case.
As shown in the figure, the controller successfully stabilized the end-effector of the mobile manipulator to the desired position.
The performance of the NMPC controller is evaluated by both position and orientation errors of the end-effector with respect to the reference pose. The positional error is measured by the Euclidean distance between the end-effector position and the reference position, while the orientation error is measured by the evaluation metric 
\begin{equation} \label{equ:metric}
E^{\theta} = 3 - \Tr(R^{I}_{E} R_{r}^{\top}),
\end{equation}
where the trace of the error rotation matrix is used. In essence, Eq.~\eqref{equ:metric} indicates that the orientation error $E^{\theta}$ converges to zero as the the end-effector orientation $R^{I}_{E}$ converges to the reference orientation $R_{r}$. 

\begin{figure}[tp]
	\centering
	\includegraphics[trim=20mm 15mm 20mm 10mm,scale=0.335]{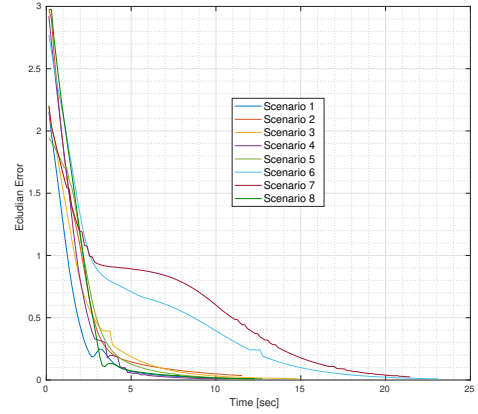}
	\caption{\footnotesize Positional error (in meters) of the end-effector.}
	\label{fig:trans_error}
\end{figure}

\begin{figure}[tp]
	\centering
	\includegraphics[trim=20mm 15mm 20mm 5mm,scale=0.335]{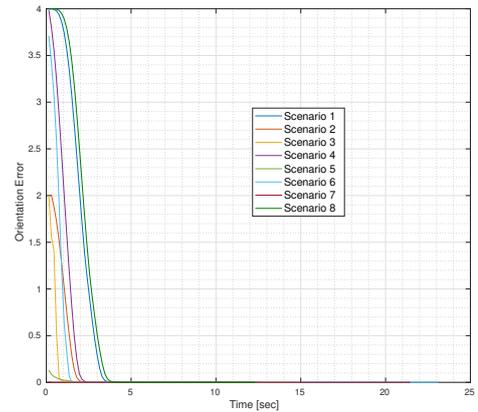}
	\caption{\footnotesize Orientation error of the end-effector for all scenarios.}
	\label{fig:ori_error}
\end{figure}

As shown in Fig.~\ref{fig:trans_error}, the positional error converges to zero for all scenarios considered. Note that in scenarios $6$ and $7$, the settling time of the controller is relatively larger than that for the other scenarios due to the joints limits. In these two cases, the mobile manipulator had to take a longer maneuver to reach the required position and orientation without violating any joint limits. In Fig.~\ref{fig:ori_error}, the orientation error of the end-effector, i.e. $E^\theta$, is shown. The error is calculated using the evaluation metric stated in~\eqref{equ:metric}. As can be seen in the figure, the controller managed to quickly achieve all the required orientations.

All the real-time simulations were executed using an Intel Core i$7$ CPU with $2.10$ GHz processor. The average computation time of OCP~\eqref{equ:opt_problem_mul} throughout all the simulation scenarios was $61 $ ms with a maximum computation time of $130$ ms and a standard deviation of $23.9$ ms. Considering that the sampling time used is $\tau = 150$ ms, the computational results suggest that the proposed NMPC algorithm meets the real-time implementation requirements while generating feasible solutions. 


\section{Conclusion and Future Work}
\label{sec:conclusion}
In this paper, we proposed an NMPC controller for end-effector stabilization of a 10-DOF mobile manipulator. We used the kinematic models for both the mobile base and the robot arm to realize a task-space control for the end-effector of the mobile manipulator. Required constraints were directly considered, which include the joint limits and the manipulator singularity. We remark that using the developed model along with NMPC, the stabilization of the end-effector was achieved without the need of any inverse kinematics solvers. Therefore, the proposed controller is a stand-alone high level controller, which only requires the localization feedback of the mobile base and the joint positions feedback of the robotic arm to operate on the mobile manipulator.  

The controller was validated through real-time dynamic simulation scenarios. The results showed efficacy and efficiency of the proposed NMPC controller. Throughout all the designed scenarios, the controller managed to smoothly stabilize the end-effector to the required pose. 

As for the future work, experiments using the real platform in Fig.~\ref{fig:robot} will be designed and performed to further validate the controller.

\section*{Acknowledgments}
We acknowledge the support of the Natural Sciences and Engineering Research Council of Canada (NSERC), [funding reference numbers PDF-532957-2019 (M.W. Mehrez), STPGP 506987 (S. Jeon)]. 

\bibliography{ifacconf}
\end{document}